\documentclass[fleqn,10pt]{wlscirep}
\usepackage[utf8]{inputenc}
\usepackage[T1]{fontenc}
\usepackage{graphicx}
\title{Computer-Aided Fall Recognition Using a Three-Stream Spatial-Temporal GCN Model with Adaptive Feature Aggregation}

\author[1]{Jungpil Shin}
\author[1]{Abu Saleh Musa Miah}
\author[1]{Rei Egawa}
\author[1]{Koki Hirooka}
\author[2]{Md. Al Mehedi Hasan}
\author[1]{Yoichi Tomioka}
\author[3,*]{Yong Seok Hwang}

\affil[1]{School of Computer Science and Engineering, The University of Aizu, Aizuwakamatsu, Japan (e-mail:musa@u-aizu.ac.jp, m5271006@u-aizu.ac.jp, ytomioka@u-aizu.ac.jp)}
\affil[2]{Department of Computer Science \& Engineering, Rajshahi University of Engineering \& Technology, Rajshahi-6204, Bangladesh (e-mail: mehedi\_ru@yahoo.com)}
\affil[3]{Electronics engineering, Kwangwoon University (e-mail: thestone@kw.ac.kr)}
\affil[*]{Corresponding author: Jungpil Shin (e-mail: jpshin@u-aizu.ac.jp), Yong Seok Hwang (e-mail: thestone@kw.ac.kr)}

\keywords{human activity recognition (HAR), AlphaPose, Fall detection (FD), Graph convolutional network (GCN), Spatial-temporal, Sep-TCN, Multi-stream deep learning,  Ageing people, Classification.}

\begin{abstract}
The prevention of falls is paramount in modern healthcare, particularly for the elderly, as falls can lead to severe injuries or even fatalities. Additionally, the growing incidence of falls among the elderly, coupled with the urgent need to prevent suicide attempts resulting from medication overdose, underscores the critical importance of accurate and efficient fall detection methods. In this scenario, a computer-aided fall detection system is inevitable to save elderly people's lives worldwide. Many researchers have been working to develop fall detection systems. However, the existing fall detection systems often struggle with issues such as unsatisfactory performance accuracy, limited robustness, high computational complexity, and sensitivity to environmental factors due to a lack of effective features.  
In response to these challenges, this paper proposes a novel three-stream spatial-temporal feature-based fall detection system. Our system incorporates joint skeleton-based spatial and temporal Graph Convolutional Network (GCN) features, joint motion-based spatial and temporal GCN features, and residual connections-based features. Each stream employs adaptive graph-based feature aggregation and consecutive separable convolutional neural networks (Sep-TCN), significantly reducing computational complexity and model parameters compared to prior systems. Experimental results across multiple datasets demonstrate the superior effectiveness and efficiency of our proposed system, with accuracies of 99.51\%, 99.15\%,  99.79\%  and 99.85 \% achieved on the ImViA, UR-Fall, Fall-UP and FU-Kinect datasets, respectively. The remarkable performance of our system highlights its superiority, efficiency, and generalizability in real-world fall detection scenarios, offering significant advancements in healthcare and societal well-being.
\end{abstract}
\begin{document}

\flushbottom
\maketitle
%
%
\thispagestyle{empty}


\section {Introduction}
\label{sec1}
The rising number of elderly individuals worldwide, expected to double over the next three decades, has highlighted the importance of fall detection technologies.
When someone is unable to react to stimuli and loses awareness of their surroundings, they may become unconscious, resembling a state of sleep and potentially drifting into slumber. This susceptibility to falls is a major worry for older individuals in Japan, resulting in both injuries and fatalities~\cite{lord2006visual}. Falls rank as the second leading cause of accidental death and disproportionately affect individuals aged 60 and above, reported to the World Health Organization (WHO)~\cite{zahedian2021effect}. Japan, known for its rapidly ageing population, is particularly concerned about the high incidence of falls among its elderly citizens. The increasing number of elderly people worldwide is a significant trend, with projections indicating that the global population of seniors will more than double over the next three decades. By 2050, it is estimated that approximately 16.0 per cent of the world's population will be elderly~\cite{united2021world}. Prompt treatment and intervention are crucial when individuals lose consciousness; otherwise, they face significant risks. 
In this scenario, It's vital to develop an automatic fall detection method to protect lives in such situations. Some falls are severe and require immediate medical attention. 
Moreover, fall detection systems have diverse applications, spanning assisted living facilities, hospitals, and homes. They offer continuous monitoring and rapid assistance in the event of a fall, ensuring the safety and well-being of individuals in these environments.

Research has shown that receiving prompt medical attention after a fall can reduce the risk of mortality by 80\% and the need for extended hospitalization by 26\%~\cite{romeo2020image}. Wearable sensors and vision-based data modalities \cite{hassan2024deep_har_miah,10624624_lstm,miah2024hand_multiculture} are the main types of the current fall detection research data collection approach \cite{gutierrez2021comprehensive}.  Sensor dataset modalities approach needs to wear the sensor and that can capture and detect the abrupt change of acceleration because of the fall \cite{lu2018image}. However, it makes some portability and flexibility complexities for elderly patients besides the drawback of the false detection because of the lying prone or setting stages~\cite{huang2018video}.
In recent times, cameras have become increasingly prevalent in both public spaces like train stations and bus stops, as well as private settings such as office buildings. This surge in camera usage aligns with the rapid advancements in computer vision, particularly driven by deep learning techniques~\cite{miah2023dynamic_mcsoc,miah2023skeleton_euvip,shin2024japanese_jsl1,kakizaki2024dynamic_jsl2,dong_deep}. These advancements have greatly aided in the development of vision-based fall detection methods~\cite{miah2023multistage,miah2022bensignnet,shin2023rotation,shin2023korean,rahim2020hand}. However, using video feeds directly in research may violate individual privacy and lead to legal complications. As a result, some studies have employed encoding techniques to obscure the clarity of the video, but they still face difficulties in ensuring the privacy of the personal identity. 

Recently, researchers utilized joint skeleton points of the body instead of the RGB-based image or video input data modalities for human action recognition, which was achieved with pose estimation algorithms \cite{vakunov2020mediapipe} and Kinect systems \cite{liu2020disentangling,keskes2021vision}. The advantage of skeleton-based data modalities is that they disappear important information to hide the personal identity and security of the patient. One of the pose-based skeleton point's main strengths is that retrieving video using reverse engineering is impossible, which leads to becoming the most suitable data modality without compromising patient privacy.  In addition, it is also robust against redundant background, lighting conditions and partial occlusion of other noises. Moreover, the joint skeleton is lower than the RGB video dimension. It can give us effective motion information and patterns over time that generate complex relationships among the joints with an acceptable computation complexity.  

Recent research has increasingly utilized pose estimation algorithms \cite{wang2020fall,xu2020fall} to leverage low-cost cameras. These algorithms directly extract 2D skeletons from video frames, ensuring both privacy and cost-effectiveness.
Recently, some researchers developed a 2D skeleton-based fall detection system using a CNN module. This system also aggregated the features using generalized filters in terms of filter size and number. 
Recently, Sania et al. \cite{zahan2022sdfa}. employed a GCN with a separable convolutional neural network in two streams, the skeleton and motion streams, to enhance the information. Where they used the CNN module for the skeleton and motion information and concatenated CNN features fed into the GCN Sep-TCN and GCN module, they used the URFD and UPFD datasets to evaluate their model, where they reported more than 94\% accuracy on average for two benchmark fall detection datasets. Also, their model seems effective, but they still face challenges in achieving good performance accuracy. To increase the performance accuracy, Egawa et al. employed a GCN-based spatial-temporal feature extraction and classification model for fall detection \cite{electronics12153234_egawa}. Then, they used the ImVia RU-Fall detection dataset to evaluate their model, where they achieved high-performance accuracy, which is around 99.00\% in both datasets. The problem with the work is that they only used motion information as input for their model. The drawback is that they used only motion-based features that seem to face difficulty achieving good performance because of their effectiveness. Sania et al. also employed skeleton and motion two streams, but they concatenated early and did not extract Sep-TCN and GCN features for both streams. In addition, they have an explanation for lost information because of the early fusion \cite{zahan2022sdfa}. 
To overcome the challenges, we propose a three-stream spatial-temporal feature-based fall detection system incorporating joint skeleton-based spatial and temporal Graph Convolutional Networks (GCN) with the late stream to overcome the issues. 

The primary contributions of this study are detailed~below:

\textbf{Novel Architecture:} We proposed an effective fall detection system that utilizes skeletons extracted from standard videos, enabling the use of low-specification cameras for surveillance. We investigated various resolutions to confirm our system's resilience against video quality. Experimental results confirm the system's effectiveness across multiple datasets with varying frame resolutions and assessment configurations. In the proposed system, we employed three streams to extract effective features. The first stream is constructed with joint skeleton-based spatial and temporal GCN features, the second with joint motion-based spatial and temporal GCN features, and the third with the residual connection.

\textbf{Strategy of the Proposed Multi-stream Concept:} The first stream processes joint skeleton information through GSTCN modules, capturing spatial and temporal features by utilizing spatial graph convolution and Sep-TCN. The second stream processes joint motion information similarly, while the third stream employs residual connections to mitigate information loss. Features from all three streams are concatenated to produce a comprehensive feature set, which is then fed into the classification module to predict fall events.

\textbf{Computational Cost Effectiveness:}
In each stream, we suggest adaptive graph-based feature integration with two consecutive distinct convolutional neural networks to capture gradual and rapid temporal movements. Then, we concatenated the features to generate the final features. The Sep-TCN in each stream helps to reduce the floating point operations per second (FLOPS) and the parameter count of our model compared to existing systems.

\textbf{Generalization:} Finally, we feed the concatenated features into the classification module to generate a probabilistic map and prediction result. After extensive experimentation, our proposed model achieved 99.51\%, 99.15\%, 99.79\%, and 99.85\% accuracy for the ImViA, UR-Fall, Fall-UP, and FU-Kinect datasets, respectively. The exceptional accuracy achieved with these datasets demonstrates the superior effectiveness and efficiency of the suggested system.

The paper's remaining sections are organized as follows: Section~\ref{sec2}~provides a summary of previous research and associated challenges. In Section~\ref{sec3}, the four fall detection benchmark datasets are introduced. The design of the proposed system is detailed in Section~\ref{sec4}. Evaluation procedures, including a comparison with a state-of-the-art approach, are elaborated on in Section~\ref{sec5}. Finally, Section~\ref{sec6}~contains our conclusions and outlines potential future research directions.

\section{Related Work} \label{sec2}
The global increase in the elderly population, projected to double over the next 30 years, has brought attention to fall detection technologies. Falls, the second leading cause of accidental death, particularly affect individuals aged 60 and above. Japan, a rapidly ageing nation, faces heightened concerns due to the prevalence of falls among its elderly population. Prompt intervention is crucial following a fall, making the development of automatic fall detection methods vital for saving lives in such scenarios \cite{united2021world,zahedian2021effect}. To Develop the system research scholars mainly use sensor-based or vision-based data modalities to develop this fall detection system. Sensor-based systems involve monitoring individuals using wearable sensors such as gyroscopes, accelerometers, EMGs, and EEGs \cite{miah2020motor,miah2019eeg}. 
These sensors capture diverse signals; various features are extracted, including angles, distances, and statistical measures \cite{mubashir2013survey}. For instance, Wang et al. utilized accelerometer data to calculate the SVMA, achieving an accuracy of 97.5\% \cite{wang2014enhanced}. Desai et al. employed multiple sensors and logistic regression to trigger emergency notifications upon detecting falls \cite{desai2020novel}. Also, Sensor-based systems provide real-time monitoring and accurate fall detection, but patients need to wear sensors continuously, and high noise levels can impact performance. Due to the mentioned limitations, researcher scholars focus on the vision-based system.  Vision-based systems with cameras have gained popularity due to their affordability and portability. These systems analyze images or videos to detect fall incidents \cite{Zerrouki_HMM,chua2015simple,cai2020vision,chen2020vision,harrou2019integrated,han2020two}. 

There are many research scholars who employed handcrafted feature extraction and machine learning-based classification methodologies \cite{kwolek2014human,youssfi2021fall,charfi2013optimised,mubashir2013survey,8741603,9230773,Zerrouki_HMM,chua2015simple}.

Also, machine learning algorithms can achieve good performance accuracy in some cases, but they face difficulties in achieving good performance accuracy and efficiency with large-scale or dynamic video datasets. To overcome the challenges, some researchers employed a deep learning-based approach \cite{cai2020vision,chen2020vision,harrou2019integrated,han2020two}. Also, the handcrafted have been used to develop performance accuracy, but deep learning shows more impact compared to deep learning \cite{le2014analysis}. 
Tian et al. applied a deep learning model and achieved 92.00\% and 94.00\% as precision and sensitivity, respectively  \cite{tian2018rf}. To enhance the performance accuracy and efficiency, Chen et al. employed the R-CNN approach to extract the effective features and then fed the features into the bi-directional LSM as a classification module ~\cite{chen2020vision}. Upon assessing their model using the UR-Fall detection dataset, they achieved a notable accuracy of 96.70\%, surpassing the performance of the prior module. 
To expand the system, Harrou et al. applied a multi-step process including preprocessing, segmentation into five different regions and feature extraction from all regions, and then they computed generalized likelihood ratio (GLR)~\cite{harrou2019integrated}. After evaluating their model with URFD and FDD datasets, they reported 96.66\% and 96.84\% accuracy, respectively. 

Han et al. aimed to enhance the efficiency of the fall detection system while maintaining stable performance accuracy. They employed a Mobile-VGG network to extract features and conduct classifications, focusing on extracting motion characteristics from RGB video sources  ~\cite{han2020two}. Conventional camera and image-based systems can face challenges in robustness, with limitations in effectively distinguishing between foreground and redundant background, potentially impacting their overall effectiveness.
To reduce the computational complexity, more recently, some researchers employed skeleton-based, which offers robustness to scene variations, redundant background and lighting conditions \cite{yan2018spatial,yao2019improved,tsai2019implementation,zheng2022lightweight}. These approaches involve extracting characteristics from skeleton joints and applying machine learning algorithms for classification \cite{yan2018spatial,yao2019improved,tsai2019implementation,zheng2022lightweight}. To enhance the fall detection system, Tran et al. Applied floor plane-based feature extraction and classification approach.  In addition, they also applied velocity magnitude between the spine and the head on the floor. Finally, they applied SVM for the classification. Yao et al. enhanced performance accuracy by extracting features from the body skeleton points. Their model was evaluated using the TSTV2 dataset, achieving an accuracy of 93.56\% with the SVM module \cite{yao2019improved}.

This article's feature extraction is to divide the skeleton information into five different regions and then extract features from each region, including the head, spine,  neck and centre of the spine. Tsai et al. proposed a fall detection system by extracting features from the selected joint skeleton key points, and then they fed the features into the 1DCNN for the classification~\cite{tsai2019implementation}. However, research scholars found some challenges, including reference interference and detection, because of the feature extraction and classification \cite{le2014analysis}.
Addressing the challenges of feature selection and ensuring optimal performance in multi-class classification scenarios remain areas of focus. While machine learning algorithms show promise, the selection of relevant features and managing computational complexity pose significant challenges. Researchers are exploring deep learning techniques to address these issues and improve the accuracy and efficiency of fall detection systems~\cite{rubenstein2006falls,chen2020vision,gasparrini2016proposal}. 

The challenge with these features lies in the fundamental difference between skeleton data, represented as a graph, and images or videos, which typically consist of 2D or 3D grids. This distinction makes it difficult for traditional feature extraction methods to accurately process the information and manage the data structure effectively without preprocessing. Many existing approaches address this issue by converting the joint points into a more standardized data format metric. Data loss during transformation can erase crucial joint connections, emphasizing the importance of data accuracy to preserve valuable insights and implementing robust validation measures to prevent such losses. Yan et al. introduced a novel deep learning strategy known as the spatial-temporal graph convolutional network (ST-GCN) to enhance performance accuracy and efficiency~\cite{yan2018spatial}. The primary focus lies in extracting diverse node connections, specifically spatial and temporal contextual links among joints, utilizing a graph-based approach rather than traditional 2D grids, resulting in the successful identification of hand motion and activities. Keskes et al. utilized the ST-GCN fall detection method to tackle multiple challenges within the field as documented in their work \cite{keskes2021vision}. Recent research has increasingly utilized pose estimation algorithms \cite{wang2020fall,xu2020fall} to leverage low-cost cameras. These algorithms directly extract 2D skeletons from video frames, ensuring both privacy and cost-effectiveness.
Recently, some researchers developed a 2D skeleton-based fall detection system using a CNN module. This system also aggregated the features using generalized filters in terms of filter size and number. 

Recently, Sania et al. \cite{zahan2022sdfa}. employed a GCN with a separable convolutional neural network in two streams, the skeleton and motion streams, to enhance the information. Where they used the CNN module for the skeleton and motion information and concatenated CNN features fed into the GCN Sep-TCN and GCN module, they used the URFD and UPFD datasets to evaluate their model, where they reported more than 94\% accuracy on average for two benchmark fall detection datasets. Also, their model seems effective, but they still face challenges in achieving good performance accuracy. To increase the performance accuracy, Egawa et al. employed a GCN-based spatial-temporal feature extraction and classification model for fall detection \cite{electronics12153234_egawa}. Then, they used the ImVia RU-Fall detection dataset to evaluate their model, where they achieved high-performance accuracy, which is around 99.00\% in both datasets. The problem with the work is that they only used motion information as input for their model. The drawback is that they used only motion-based features that seem to face difficulty in achieving good performance because of the feature's effectiveness. Sania et al.\cite{zahan2022sdfa}. They also employed skeleton and motion in two streams, but they concatenated early and did not extract Sep-TCN and GCN features for both streams. In addition, they have an explanation for lost information because of the early fusion. To overcome the challenges, we proposed a three-stream spatial-temporal feature-based fall detection system where we incorporate joint skeleton-based spatial and temporal Graph Convolutional Networks (GCN).


\section{Datasets} \label{sec3}
The availability of online dynamic fall detection benchmark datasets is scarce, prompting the selection of three specific benchmark datasets for this research: the Fall-UP Dataset \cite{HAR_Dataset}, ImViA Datasets (le2i) \cite{charfi2013optimised}, UR Fall Dataset~\cite{kwolek2014human} and FU-Kinect\cite{aslan2017skeleton_FU-kinetic}. Table~\ref{Tab:Dataset_Table} provides an overview of datasets used in this research, including their specifications, features, individuals, and actions, among other details. The Fall-UP Dataset includes 11 classes with 62K images, ImViA Datasets (le2i) features two classes with 40K images, and the UR Fall Detection dataset comprises two classes with 3K images, FU-Kinect features six classes with 120.9k images in total.

\begin{table}[ht]
\centering
\caption{Overview of the datasets utilized in this research.}
\label{Tab:Dataset_Table}
\setlength{\tabcolsep}{5pt}
\begin{tabular}{|l|l|l|l|}
\hline
\textbf{Name of the Dataset} & \textbf{Tpe} & \textbf{No. Classes} & \textbf{No. Sample} \\
\hline
 \begin{tabular}[c]{@{}l@{}}Fall-UP Dataset\\ \cite{HAR_Dataset}\end{tabular} & \begin{tabular}[c]{@{}l@{}}Raw vidoes.\end{tabular} & \begin{tabular}[c]{@{}l@{}}11 Class\end{tabular} & \begin{tabular}[c]{@{}l@{}}62 K images\\  in total\end{tabular} \\
 \begin{tabular}[c]{@{}l@{}}ImViA Datasets\\ (le2i) \cite{charfi2013optimised}\end{tabular} & \begin{tabular}[c]{@{}l@{}}Raw vidoes\end{tabular} & \begin{tabular}[c]{@{}l@{}}2 Class Fall/ \\ Non-Fall\end{tabular} & \begin{tabular}[c]{@{}l@{}}40 K images\\  in total\end{tabular} \\
\begin{tabular}[c]{@{}l@{}}UR \\ Fall Detection~\cite{kwolek2014human}\end{tabular} & \begin{tabular}[c]{@{}l@{}}
Raw vidoes\end{tabular} & \begin{tabular}[c]{@{}l@{}}2 Class Fall/\\  Non-Fall\end{tabular} & \begin{tabular}[c]{@{}l@{}}3 K images \\ in total\end{tabular} \\
\begin{tabular}[c]{@{}l@{}}FU- \\ Kinetic~\cite{wang2014cross}\end{tabular} & \begin{tabular}[c]{@{}l@{}} Raw vidoes\end{tabular} & 6 class & \begin{tabular}[c]{@{}l@{}}120.9 K images \\ in total\end{tabular} \\
\hline
\end{tabular}
\end{table}

\subsection{Fall-UP Dataset}
Fall-UP is another benchmark fall detection dataset constructed with 11 different activities, each recording 3 trials. This dataset was recorded from 17 people, 9 of whom were males and the remaining 8 females, who were 18-24 years old. Their average height was 1.66m, and their average weight was 66.6kg. Of the 11 activities, six were normal daily activities, and 5 were falls. 
 Daily activities lasted 60 seconds, except jumping (30 seconds) and picking up an object (10 seconds)~\cite{HAR_Dataset}.

\subsection{UR Fall Detection~Datasets}
The UR Fall Detection Dataset~\cite{pathak2015fall} is one of the most benchmark fall datasets in this domain, which consists of short clips of videos with both falling and non-falling scenarios. This dataset is recorded by the Rehabilitation Medicine Research Group at the University of Rochester, and it contains a total of 30 fall dataset videos ~\cite{kwolek2014human}. The dataset was recorded with two separate cameras, resulting in a total of 70 active cameras and 3000 images, and it contained 30 fall videos and 40 non-fall videos with daily activities. As we said, this dataset mainly comprises RGB and Depth camera recordings at 640x480 resolution, along with annotations detailing fall event timing and individual postures. This dataset serves as a valuable resource for fall detection research and is widely used as a benchmark in various studies. Table \ref{Tab:Dataset_Table} describes the basic information of this dataset.

\subsection{ImViA Datasets(le2i) }
ImViA is another benchmark dataset for fall detection, comprising videos recorded by a single camera in a realistic surveillance environment~\cite{charfi2013optimised}. It is constructed with a range of daily activities like moving between furniture, exercising, and falling. Each video features a single individual, with a resolution of 320 $\times$ 240 pixels and a frame rate of 25. The background remains consistent, while the textures in the images are complex.

\subsection{FU-Kinect Dataset}
The FU-Kinect-Fall dataset was developed using the Kinect V1 device. It includes actions such as walking, bending, sitting, squatting, lying, and falling, performed by 21 individuals aged between 19 and 72 years. In total, the dataset contains 1008 depth videos and 3D coordinates $(x, y, z)$ of 20 joints, with each action repeated 8 times by each subject, resulting in $6 \times 8 \times 21$ recordings. Each video lasts approximately 4-5 seconds, recorded at a resolution of $320 \times 240$ and 30 frames per second, varying slightly based on the action.
To create the dataset, a Kinect V1 camera was mounted on a tripod at a height of 95 cm from the ground. The camera covered a reference range of 0.5 m to 4.5 m, within the Kinect's visual limits. The actions were performed in a $4 \times 4 \, \text{m}$ area, with the depth sensor positioned vertically, and the subjects were situated between 2 m and 3.5 m from the camera. The dataset also took into account the physical characteristics of the subjects, such as height and weight, and the specific actions they performed \cite{aslan2017skeleton_FU-kinetic}. The dataset is available at \url{https://github.com/MuzafferAslan23/Fall-Detection-Dataset}.


\section {Proposed~Methodology} \label{sec4}


The methodology employed in this study aims to develop an efficient fall detection system that addresses the limitations of existing methods. We begin by collecting standard videos containing human activities, focusing particularly on scenarios relevant to fall detection. These videos serve as the primary data source for training and evaluating our proposed model. Figure \ref{Figure:Main_Figure} demonstrates the working flowograph of the proposed model. Firstly, we extracted the information on the human pose joint skeleton from the video that captured human subjects' spatial configuration and motion dynamics. This step involves employing state-of-the-art computer vision techniques to accurately detect and track key body joints in each video sequence frame. 

\begin{figure}[htp]
\centering
\includegraphics[width=14 cm]{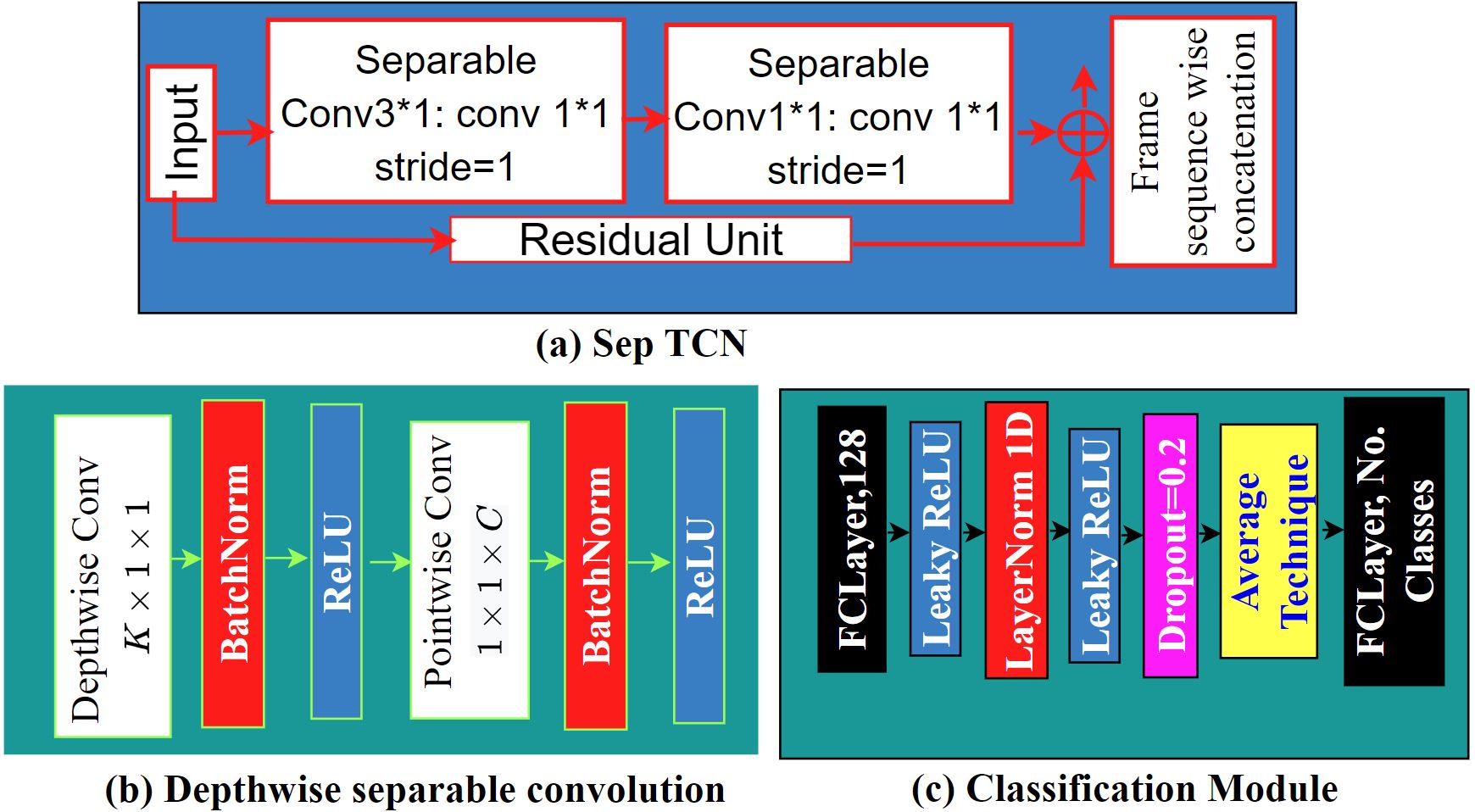}
\caption{{Details of each module (a) Sep TCN (b) Depthwise separable convolution (c) Classification Module}}
\label{Figure:Main_Figure_details}
\end{figure}

Then, we employed a novel architecture for fall detection, consisting of multiple streams for feature extraction. Specifically, we integrate joint skeleton-based spatial and temporal Graph Convolutional Network (GCN) features, joint motion-based spatial and temporal GCN features, and residual connections to enhance feature representation and capture relevant patterns in the data. Furthermore, we introduce adaptive graph-based feature aggregation within each stream and consecutive separable convolutional neural networks (Sep-TCN) \cite{miah2024sign_largescale,miah2024spatial_paa,}. These techniques reduce computational complexity and model parameters while preserving discriminative information crucial for accurate fall detection. Figure \ref{Figure:Main_Figure_details} demonstrates the details of each module of the proposed model. 

According to Figure \ref{Figure:Main_Figure}, the proposed model is constructed with three streams: stream-1: spatial-temporal features from joint skeleton information, stream-2: spatial-temporal features from joint-motion information, and stream-3: residual unit or skip connection to overcome information loss issues. The first and second streams are constructed with consecutive GSTCN, which are composed of spatial graph convolution (SGC) and Sep-TCN, where SGC is employed to enhance spatial contextual features, and Sep-TCN is employed to enhance temporal contextual features obtained from the SGC.

The first stream takes the joint skeleton as input and feeds it into the GSTCN-11, which is constructed with SGC and Sep-TCN. The idea is that it can produce spatial enhancement features with SGC and then enhance the temporal features with Sep-TCN. To achieve long-range dependency, we consecutively fed the output of GSTCN-11 into the GSTCN-12 module. In the second branch, we computed the motion from the joint skeleton information and then fed it into GSTCN to enhance the spatial-temporal features and achieve a long-range dependency feature with GSTCN-22. The first and second streams produce spatial-temporal contextual features from the joint skeleton information and joint motion information, respectively. To overcome information loss issues, we used a skip connection as a third stream. Finally, we concatenated the three stream features and produced the final feature that was fed into the classification module. 

To evaluate the performance of our proposed system, we conduct extensive experiments using multiple datasets, including ImViA, UR-Fall, Fall-UP and FU-Kinect. We measure the system's effectiveness by calculating accuracy metrics and comparing its performance against baseline methods and state-of-the-art approaches.
Through rigorous experimentation and analysis, we demonstrate our proposed system's superior effectiveness and efficiency in real-world fall detection scenarios. The results validate the efficacy of our methodology and highlight its potential for significant advancements in healthcare and societal well-being.

{\begin{figure}[ht] 
\centering
\includegraphics[scale=0.25]{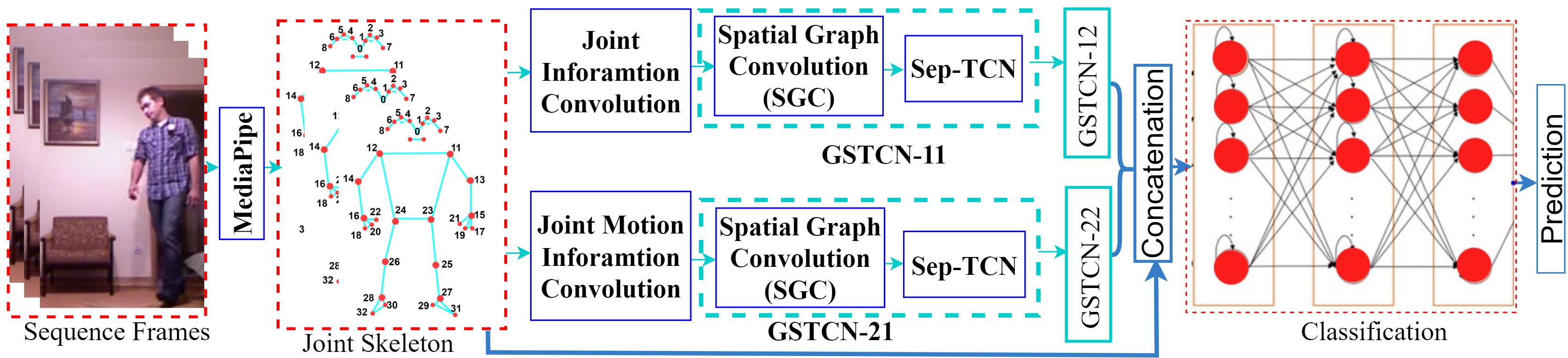}
\caption{Proposed working flow diagram.}
\label{Figure:Main_Figure}
\end{figure}}

\vspace{6pt}

\subsection{Pose~Estimation}
\begin{figure}[htp]
\centering
\includegraphics[scale=0.6]{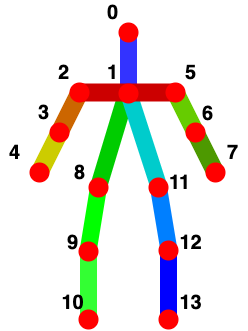}
\caption{{Visual representation of body skeleton joints produced using AlphaPose.}}
\label{Figure:alphapose}
\end{figure}

The joint skeleton keypoint reduces the dimension of the dataset and secures the patient's privacy information. To extract key points based on the pose extraction method, media pipe \cite{vakunov2020mediapipe} is one of the used approaches. However, in the study, we used the Alphapose approach to extract the joint skeleton from the RGB video input dataset.  The AlphaPose, a free-source visual image processing library, extracts skeletal joints from our fall detection dataset. It employs a deep learning-driven pre-trained model, enabling real-time applications like facial recognition, object detection, and pose estimation. AlphaPose utilizes a top-down approach, ensuring precise joint point detection for each individual. We discarded frames where AlphaPose failed to extract skeletons, focusing only on successful extractions~\cite{fang2017rmpe,xiu2018pose,fang2022alphapose}.
After extracting the joint skeleton using the Alphapose model, we included the detailed information in Table~\ref{Tab:Alpha_getframe}, where we also mentioned that Alphapose sometimes failed to extract key points for some frames. Alphapose generated 18 joint skeleton key points, which we demonstrated in Figure \ref{Figure:alphapose}. 
This approach can be processed offline or online to extract the corresponding joint skeleton key points. The 18 points it produced from each frame, namely, nose, left -right shoulders, left -right elbows, left -right hips, and left -right knees.

\begin{table*}[htp]
\centering
\caption{The number of frames successfully extracted skeletons using AlphaPose.}
\label{Tab:Alpha_getframe}
\begin{tabular}{|l|l|l|l|}
\hline
\textbf{Name of the Dataset}&\textbf{Total Frames}&\begin{tabular}[l]{@{}l@{}}\textbf{Number of Frames } \\ \textbf{for Which the Skeleton}\\ \textbf{Could Be Acquired}\end{tabular} &\begin{tabular}[l]{@{}l@{}}\textbf{Number of Frames}  \\ \textbf{for Which the Skeleton}\\ \textbf{Could Not Be Acquired}\end{tabular}\\ \hline
ImViA&42,066&40,631&1,435\\ 
UR fall dataset&2,995&2,142&853\\ 
Fall-Up Dataset & 628,036&554,857&73,179\\
\hline
\end{tabular}
\end{table*}

\subsection{Motion Calculation and Graph~Construction}
\begin{figure*}[htp]
 \centering
\includegraphics[scale=0.25]{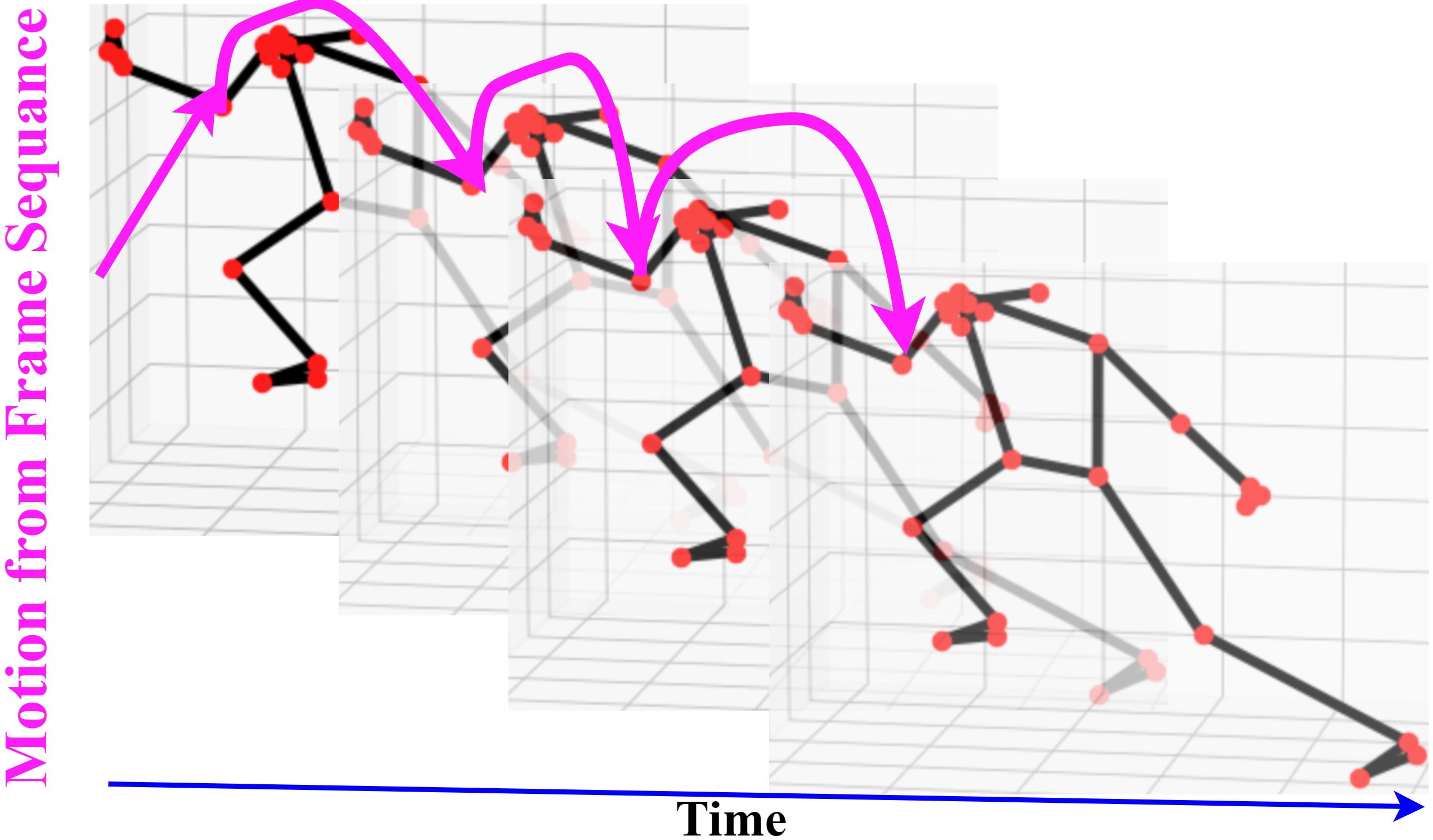}
\caption{Example illustrating the motion calculation process.}
\label{Figure:motion_calculation}
\end{figure*}

Skeleton joints provide valuable information about spatial and temporal connections. These links stay constant over time and any changes in how they are arranged in space and time act as unique indicators of various activities. However, certain activities like squatting, exercising or picking up something from the ground, falling, and lying down share similar posture transitions. The key difference lies in their speed. Falls, for instance, involve rapid changes in joint coordinates within a narrow timeframe compared to other actions. Thus, the proposed framework incorporates motion trajectory to enhance the distinction between these different action types. In this study, we focused on analyzing a dynamic fall detection dataset. Motion stands out as a crucial feature for dynamic fall detection, influencing movement, alignment, and overall data structure effectiveness. To capture motion, we computed it using all landmarks as two-dimensional vectors (\textit{X} and \textit{Y}). We calculated the motion for each joint by comparing the positions of joints in consecutive frames. This procedure is depicted in Figure~\ref{Figure:motion_calculation}.
The motion (\textit{M}) for a particular joint (\textit{j}) was computed by subtracting consecutive joint positions in frames, as specified in Equation~(\ref{Eq:motion}).

\begin{equation} \label{Eq:motion} 
M(j) = 
\begin{cases}
ActivityMotion_{X}=X_{t}-X_{t-1}  \\
ActivityMotion_{Y}=Y_{t}-Y_{t-1}.
\end{cases}
 \end{equation}

After extracting the joint and motion information, we constructed a graph using that information. The main goal of the graph construction is to extract spatial and temporal contextual enhancement. It is based on the natural connection of the human body or the random connection among joints based on the provided structures. We constructed the graph with the below Equation (\ref{Eq:graph_construction}).

 \begin{equation} \label{Eq:graph_construction} 
G=(V,E)                           
\end{equation}
To denote the set of nodes and edges, we used $V$ and $E$, respectively, in Equation~(\ref{Eq:graph_construction}). The graph based on the whole body skeleton can be denoted as $V = \{ v(i,t) \mid i=1, \ldots, N, \, t=1, \ldots, T \}$. Based on the internal connection structure, we formed an adjacency matrix using the following formulas in Equation~(\ref{Eq:adj_mtx}):

\begin{equation} \label{Eq:adj_mtx}
f(x) =
\begin{cases}
1 & \text{if they are adjacent} \\
0 & \text{if they are not adjacent}.
\end{cases}
\end{equation}
 
\subsection{Spatial Graph Convolutional~Network }
The Spatial Graph Convolutional Network captures spatial correlation among joints. 
For a given input skeleton, it consolidates pertinent topological connections from geometrically significant subgraphs. Convolution operates under the guidance of an adjacency matrix $A$, which depicts the connectivity of the skeleton graph among joints and bones. Convolutional modelling is conducted separately over a joint set B, encompassing both the present joint and its immediate neighbours, to extract relevant relationships. 
Our goal is to extend the joint connection where the matrix can be dynamically adjusted with the input data. To do this, we introduce a trainable parameter M to refine the adjacency matrix with the increasing number of channels in each layer \cite{Zahan@SDFA}. The spatial convolutional model according to our concept can be written as the below Equation \ref{eq:spatial_convolutin}.

\begin{equation} \label{eq:spatial_convolutin}
f_{1}(v_{ti}) = \sum _{v_{tj}\in N(v_{ti})}\frac{1}{Z_{ti}(v_{tj})}\mathcal {F}\left(X(v_{tj})\right)W(l(v_{tj})) \tag{4} 
\end{equation}

where $v_{tj}$ denotes the $jth$ dynamic motion-imbued joint at time $t$, $N(v_{ti})$ signifies the collection of joints, $Z_{ti}(v_{tj})$ stands for the normalization factor. 
Then for any joint $v_{tj}$ in the layer $l$ we can write the weight matrix as $W(l(v_{tj}))$. Then signification of the skeleton frames is denoted with $X(v_{tj})$, and the embedding of the skeleton frames is denoted with $\mathcal {F}(X(v_{tj}))$ for the joint and joint motion information. Lastly, the learned adjacency matrix with neighbouring joint matrix accumulation is as the Equation \ref{eq:spatial_convolutin_1}:

\begin{equation} 
\label{eq:spatial_convolutin_1}
f_{\text{out}}(v_{ti}) = \sum _{v_{tj}\in B(v_{ti})}A(v_{tj})M_{l}(v_{tj})f_{1}(v_{tj}) \tag{5} \end{equation}
In the Equation set of the adjacent demonstrated with $B(v_{ti})$, the significance of the adjacency matrix is denoted with the  $A(v_{tj})$, learned matrix for a specific layer $l$ is denoted with the $M_{l}(v_{tj})$. After that, the $f_{1}(v_{tj})$ is the output of the  Equation (\ref{eq:spatial_convolutin}) which is generated from an specific joint  $v_{tj}$.

The Spatial graph convolution (SGC)  employed weight sharing among neighbouring joints to capture intricate spatial dependencies. Following the GSTCN, we applied a residual connection where we used  $1 \times 1$ convolution for preserving the input feature vector. Then, we utilized max pooling to enhance the spatial features, which also gave us the hierarchical feature that helps to strengthen the most active joints skeleton keypoint that leads to enhancing the spatial contextual information.

\subsection{Sep-TCN: Separable Temporal Convolutional Network}
Sep-TCN is slightly different from the conventional convolutional operation, where conventional convolutional operation is constructed with the $K \times K \times C$ mathematical operation, which leads to a complex multiplications formula that leads to high computational complexity time during the kernel shifts. This leads to an increase in the total number of FLOPS requirements. To tackle the problems, the researchers proposed Sep-TCN, which is composed of depthwise (DW) and pointwise (PW) convolutions. This module's working procedure involves dividing the calculation mentioned into two parts. 
Depthwise separable convolution divides the traditional convolution operation into two distinct phases: depthwise (DW) and pointwise (PW) convolutions. The Sep-TCN diagram is demonstrated in Figure \ref{Figure:Main_Figure_details}(a).

The first phase is the DW phase, which is constructed with $K \times K \times 1$ filter computation and produces the output based on the depth-wise dimension of the input dataset. This approach significantly reduces the number of multiplications required during the process, leading to a substantial decrease in computational cost because of the low parameters. By employing DW convolution, the model can extract spatial features efficiently without compromising performance \cite{guo2014online}.

On the other hand, pointwise convolution involves applying a $1 \times 1 \times C$ filter across the output of the depthwise convolution. This step aims to combine the spatial information extracted by the DW convolution with cross-channel interactions. By integrating the pointwise convolution, the model can learn complex relationships between different feature channels, enhancing its representational capacity \cite{guo2014online}. 

In our implementation, we utilized a kernel size of $3 \times 1$ for the first Sep-TCN layer and $1 \times 1$ for the second, resulting in filter sizes of $3 \times 1 \times 1$ and $1 \times 1 \times C$ for the depthwise convolution (DW) layer, as illustrated in Figure \ref{Figure:Main_Figure_details}(b). These configurations allowed our model to efficiently capture both local and global temporal dependencies while preserving the original joint dimension throughout the computation. Figure \ref{Figure:Main_Figure_details}(b) describes in detail the internal mechanism architecture of the DW and PW, along with a basic Sep-TCN layer. In addition, we also included the residual connection to enhance the temporal contextual information, which is constructed with max-pooling and can highlight the most crucial frames in the feature vectors. Furthermore, to enhance temporal focus, we incorporated a residual connection using temporal max pooling, which emphasizes the most important frames in the feature vector.

\subsection{Masking Operation in GSTCN Block}
To mitigate model overfitting, we implemented a technique involving randomized masking on joints and frames, encouraging our model to learn from sparsely populated feature matrices \cite{Zahan@SDFA,miah2023dynamic_graph_general,10360810_miah_ksl2,shin2024korean_ksl0}. DropGraph mechanism also implemented the concept of the making operation, which mainly integrated the dropout layer aiming to features akin. Moreover, it excludes the multiple joints for spatial cases and the sequence of the consecutive frames for the temporal cases. It seems to be the sequence frame information, such as t and t+1, but the disjoint frames at different time intervals, such as t and t+n, contained vital information that is not identical and removing such information helps to improve the generalization property of the systems. However, the successive or neighbour frame information is minimal, and removing this information has little impact on the system.  Thus, we introduce sparsity into our input skeleton sequences by randomly masking frames and joints at various positions, utilizing a determined probability during model training. Experimentation indicates that this strategy enhances generalization by aiding in the structured learning of essential cues, resulting in improved performance outcomes. In the testing case, we do not need the making operation, and there is a setting option, we can disable this by setting 0 and using the complete dataset for evaluation.

\begin{table*}[!htp]
\centering
\caption{Computal speed comparison with the state of the art model}
\label{Tab:speed}
\begin{tabular}{|l|l|l|}
\hline
\textbf{Method Name}&\textbf{Mean} \textbf{{[}ms{]}}&\textbf{Standard Deviation}\textbf{{[}sec{]}}\\ \hline
ST-GCN \cite{electronics12153234_egawa}&5.6& 6.0869\\ \hline
GSTCAN \cite{electronics12153234_egawa}&5.4&6.0363\\ \hline
\begin{tabular}[c]{@{}c@{}}Proposed \\ Model\end{tabular} &3.2& 5.1872\\ 
\hline
\end{tabular}
\end{table*}

\begin{table*}[!htp]
\centering
\caption{precision, sensitivity, and F1-score for class-wise Fall-UP~dataset.} 
\label{Tab:Fall_up_accuracy}
\setlength{\tabcolsep}{6.75mm}
\begin{tabular}{|l|l|l|l|l|}
\hline
\textbf{Label Name} & \textbf{Accuracy {[}\%{]}} & \textbf{Precision {[}\%{]}} & \textbf{Sensitivity {[}\%{]}} & \textbf{F1-Score {[}\%{]}} \\  \hline
Falling forward using hands & n/a & 99.27 & 99.14 & 99.2\\ \hline
Falling forward using knees & n/a & 99.27 & 99.27 & 99.27 \\ \hline
Falling backwards & n/a & 99.32 & 97.32 & 98.31 \\ \hline
Falling sideward & n/a & 99.30 & 99.72 & 99.51 \\ \hline
Falling sitting in empty chair & n/a & 99.39 & 99.32 & 99.35 \\  \hline
Walking & n/a & 99.89 & 99.92 & 99.91 \\  \hline
Standing & n/a & 99.95 & 99.95 & 99.95 \\  \hline
Sitting & n/a & 99.97 & 99.94 & 99.96 \\  \hline
Picking up an object & n/a & 99.86 & 99.86 & 99.86 \\  \hline
Jumping & n/a & 100 & 99.90 & 99.95 \\  \hline
Laying & n/a & 99.36 & 99.77 & 99.57 \\  \hline
Total & 99.79& 99.6 & 99.46 & 99.53  \\  
\hline
\end{tabular}
\end{table*}

\begin{table*}[!htp]
\centering
\caption{State-of-the-art comparison of the proposed model for Fall-UP dataset.} 
\label{Tab:Fall_up_comparison}
\begin{tabular}{|l|l|l||l|l|l|l|}
\hline
\textbf{Algorithm} & \textbf{Dataset} & \begin{tabular}[c]{@{}c@{}}\textbf{Accuracy}  \textbf{{[}\%{]}}\end{tabular} & \begin{tabular}[c]{@{}c@{}}\textbf{Precision} \textbf{{[}\%{]}}\end{tabular} & \begin{tabular}[c]{@{}c@{}}\textbf{Sensitivity} \textbf{{[}\%{]}}\end{tabular} &  \begin{tabular}[c]{@{}c@{}}\textbf{F-Score}\textbf{{[}\%{]}}\end{tabular} \\
\hline
Martínez~\cite{HAR_Dataset} & Fall-UP dataset & 95.00 &77.7  &69.9 &  72.8\\ \hline
\begin{tabular}[c]{@{}c@{}}Proposed Model\end{tabular} & Fall-UP dataset & 99.79& 99.6 & 99.46 & 99.53 \\
\hline
\end{tabular}
\end{table*}

\subsection{Multi-Stage of GSTCN and Feature Fusion} 
In each stream, we used two stages of the GSTCN module to enhance its hierarchical features \cite{electronics12132841_miah_multistream_4}. The utilization of two consecutive GSTCN modules in our proposed framework offers several benefits. Firstly, it enables a more intricate extraction of spatiotemporal features from the joint skeleton and motion streams. Secondly, it enhances the model's capacity to capture and represent complex patterns within the input data, leading to improved performance in fall detection. After that, we produced the final feature by concatenating three feature streams: joint stream features, joint motion stream features and skip connection features.

\subsection{Classification Module}
Then, the final features go to the classification module, which is constructed with the combination of fully connected layers, ReLU activation, layer normalization, dropout, average pooling, and a softmax activation function within the classification module, offers several strengths and advantages that contribute to enhancing the accuracy of the model. Firstly, the fully connected layers allow for complex nonlinear relationships to be learned from the extracted features, enabling the model to capture intricate patterns in the data. The ReLU activation function introduces nonlinearity, aiding in overcoming the limitations of linear transformations and enabling the model to learn more complex representations. Layer normalization helps stabilize the training process by normalizing the activations within each layer, which mitigates the internal covariate shift problem and accelerates convergence. Dropout regularization prevents overfitting by randomly dropping out units during training, forcing the network to learn robust features that generalize well to unseen data. Average pooling serves to downsample the feature maps, reducing computational complexity while retaining important information. Finally, the softmax activation function produces probability distributions over the output classes, facilitating multi-class classification tasks.
By combining these layers in a sequential manner, the model gains the capability to learn intricate representations from the data, regularize itself against overfitting, stabilize training dynamics, and produce accurate predictions. This comprehensive approach to classification empowers the model to achieve higher accuracy and generalization performance on diverse datasets.

\section{Extensive Experimental Evaluation  } \label{sec5}
We conducted a series of experiments using four benchmark datasets aiming to validate the proposed system and proved the superiority and effectiveness of the proposed system. We commenced by outlining the training configuration and evaluation metrics, then proceeded to demonstrate the performance of our model across different datasets. Ultimately, we presented a comparative table showcasing the state-of-the-art results.

\subsection{Training~Setting }\label{subsec5.1}
In the study, we performed the hyperparameter set equally for all datasets during experimental evaluation. We used 0.01 for the initial learning rate and SGD optimizer with 0.9 momentum~\cite{tieleman2012rmsprop}. Our batch size was 32, and to experiment with it, we utilized a GPU machine, which consisted of GPU Geforce GTX 1080 Ti 11GB, with 64 GB of RAM, NVIDIA driver version 535, and CUDA-11.7. We trained themodel with 100 epochs. We used pytorch to implement the system \cite{paszke2019pytorch}, in addition to opencv, csv, pickle, deep learning layers, transformer~\cite{gollapudi2019learn,dozat2016incorporating}.

\begin{table*}[htp]
\centering
\caption{precision, sensitivity, and F1-score for  class wise UR fall~dataset.}
\label{Tab:UR_accuracy}

\begin{tabular}{|l|l|l|l|l|}
\hline
\textbf{Name of the labels} & \textbf{Accuracy {[}\%{]}} & \textbf{Precision {[}\%{]}} & \textbf{Sensitivity {[}\%{]}}& \textbf{F1-Score  {[}\%{]}}\\ \hline
Fall & n/a & 100 & 86.67 & 92.86 \\ \hline
NonFall & n/a & 99.1 & 100 & 99.55 \\ \hline
Average & 99.15 & 99.55 & 93.33 & 96.2  \\ 
\hline
\end{tabular}
\end{table*}

\begin{table*}[htp]
\centering
\caption{State of the art comparison of the proposed model with UR fall~dataset.} 
\label{Tab:UR_comparison}

\begin{tabular}{l|l|l|l|l|l||l|}
\hline
\textbf{Method Name} & \textbf{Dataset} & \begin{tabular}[l]{@{}l@{}}\textbf{Accuracy}\\  \textbf{{[}\%{]}}\end{tabular} & \begin{tabular}[l]{@{}l@{}}\textbf{Precision}\\ \textbf{{[}\%{]}}\end{tabular} & \begin{tabular}[l]{@{}l@{}}\textbf{Sensitivity}\\ \textbf{{[}\%{]}}\end{tabular} & \begin{tabular}[l]{@{}l@{}}\textbf{Specificity}\\ \textbf{{[}\%{]}}\end{tabular} & \begin{tabular}[l]{@{}l@{}}\textbf{F-Score}\\ \textbf{{[}\%{]}}\end{tabular} \\
\hline
HCAE~\cite{cai2020vision} & UR Fall & 90.5\% &n/a & n/a & n/a & n/a   \\ \hline
\begin{tabular}[l]{@{}l@{}}Depth \\+SVM~\cite{kwolek2014human} \end{tabular} & UR Fall  & 94.28 & n/a & n/a & n/a & n/a \\ \hline
\begin{tabular}[l]{@{}l@{}}Skeleton \\  +SVM~\cite{youssfi2021fall} \end{tabular} & UR Fall & 96.55 & n/a & n/a & n/a & n/a  \\ \hline
Harrou~\cite{harrou2019integrated} & UR Fall & 96.66 & 94 & 100 & 94.93 & 96.91 \\ \hline
\begin{tabular}[l]{@{}l@{}}Bi-Directional\\  LSTM~\cite{chen2020vision}\end{tabular} & UR Fall & 96.70\% & n/a & n/a & n/a & n/a   \\ \hline
Hontago~\cite{zheng2022lightweight} & UR Fall & 97.28 & 97.15 & 97.43 & 97.30 & 97.29 \\ \hline
Wang~\cite{wang2020fall} & UR Fall & 97.33 & 97.78 & 97.78 & 96.67 & 97.78 \\ \hline

\begin{tabular}[l]{@{}l@{}}Proposed \\ Model\end{tabular} & UR Fall &  {99.15} & 99.55 & 93.33 & n/a & 96.2\\ 
\hline
\end{tabular}
\end{table*}

\subsection{Evaluation~Matrices }
We evaluate the proposed model with a distinct dataset and produce evaluation matrices in terms of the F1-score, precision, recall, time and accuracy, and details are given here\cite{miah2022bensignnet}, 
\begin{itemize}
\item Time: We conducted a \emph{t}-test using the mean and variance information of the processing speed of the proposed model.
\item Accuracy: The accuracy metric, widely adopted by researchers, represents the percentage of correctly classified items, calculated as (TP + TN)/(TP + TN + FP + FN).
\item Recall/Sensitivity: This metric primarily reflects the true positive rate, computed as TP/(TP + FN).
\item F1-score: Calculated with TP/(TP + 1/2 (FP + FN)), and it referred to the harmonic mean between recall and precision. 
\end{itemize}

\subsection{Comparison of the Computational Time}\label{subsec5.3.1}
The inference (forwarding) mean and standard deviation of computational time per one sample of the proposed model are demonstrated in Table~\ref{Tab:speed}, which we measured here using milliseconds (ms), for the proposed model and the state of the art. ST-GCN achieves a mean speed of 5.6 ms with a Standard Deviation of 6.0869. GSTCAN follows with a mean of 5.4 ms and a Standard Deviation of 6.0363. The proposed model outperforms both, with a mean speed of 3.2 ms and a Standard Deviation of 5.1872. The result performers based on the \emph{t}-test, which rejected the hypothesis that was constructed based on the equal processing speed and proved the greater efficiency of the proposed model. 

\begin{table*}[htp]
\centering
\caption{precision, sensitivity, and F1-score for class-wise ImViA fall~dataset.} 
\label{Tab:ImViA_accuracy}
\begin{tabular}{|l|l|l|l|l|}
\hline
\textbf{Label Name} & \textbf{Accuracy {[}\%{]}} & \textbf{Precision {[}\%{]}} & \textbf{Sensitivity {[}\%{]}} & \textbf{F1-Score {[}\%{]}} \\  \hline
Fall & n/a & 97.53 & 94.61 & 96.05 \\  \hline
NonFal & n/a & 99.74 & 99.89 & 99.81 \\ \hline
Average & 99.64 & 98.64 & 97.25 & 97.93  \\ 
\hline
\end{tabular}
\end{table*}

\begin{table*}[htp]
\centering
\caption{State-of-the-art comparison of our proposed model with ImViA fall dataset} 
\label{Tab:ImViA_comparison}
\begin{tabular}{|l|l|c|c|c|c|c|}
\hline
\textbf{Algorithm} & \textbf{Dataset} & \textbf{Accuracy {[}\%{]}} & \textbf{Precision {[}\%{]}} & \textbf{Sensitivity {[}\%{]}} & \textbf{Specificity {[}\%{]}} & \textbf{F-Score {[}\%{]}} \\
\hline
Chamle~\cite{chamle2016automated} & ImViA & 79.31 & 79.41 & 83.47 & 73.07 & 81.39 \\ 
\hline
Hontago~\cite{zheng2022lightweight} & ImViA & 96.86 & 97.01 & 96.71 & 96.81 & 96.77 \\ 
\hline
Wang~\cite{wang2020fall} & ImViA & 96.91 & 97.65 & 96.51 & 97.37 & 97.08 \\ 
\hline
Proposed Model & ImViA & 99.64 & 98.64 & 97.25 & n/a & 97.93 \\ 
\hline
\end{tabular}
\end{table*}

\subsection{Performance Result of the Fall-UP Dataset and State-of-Art Comparison}\label{subsec5.3.4}
In the study, we also evaluated the effectiveness of the proposed model by evaluating it with the Fall-UP dataset, which is demonstrated in Table~\ref{Tab:Fall_up_accuracy}. Table \ref{Tab:Fall_up_accuracy} presents the label-wise performance accuracy, precision, sensitivity, and F1-score and yields remarkable results across various activity labels. Precision rates were consistently high, with values ranging from 99.27\% to 100\%, showcasing the model's ability to identify specific activities accurately. Sensitivity scores reflected the model's adeptness in detecting true positives, with values ranging from 97.32\% to 99.92\%. Notably, the model excelled in discerning nuanced movements such as falling sideward and picking up objects, achieving sensitivities above 99.7\%. F1-Scores, harmonizing precision and sensitivity, demonstrated the model's balanced performance, with values ranging from 98.31\% to 99.96\%. Overall, the model exhibited exceptional accuracy, achieving an impressive 99.79\%. These results underscore the efficacy of the proposed multi-stage deep learning framework in accurately classifying activities within the Fall-UP dataset, offering promising avenues for enhanced fall detection systems.

We also compare the proposed model with the state-of-the-art model, which is demonstrated in Table~\ref{Tab:Fall_up_comparison}.
The state-of-the-art comparison highlights the significant advancements achieved by the proposed model in fall detection using the Fall-UP dataset. Compared to the Martínez algorithm, the proposed model demonstrated substantial improvements across all metrics. With an accuracy of 99.79\%, precision of 99.60\%, sensitivity of 99.46\%, and an F-score of 99.53\%, the proposed model outperformed Martínez's accuracy of 95.00\% and F-score of 72.8\%. These results underscore the superior performance of the proposed model in accurately identifying falls, marking a notable advancement in fall detection technology. More recent work reported 95.00\% accuracy using ML-based models with this dataset, where our proposed model achieved around 6\% higher accuracy compared to their model \cite{HAR_Dataset}. 
These results underscore the superior performance of the proposed model in accurately identifying falls, marking a notable advancement in fall detection technology.

\subsection{Experimental Result with UR Fall~Dataset and State-of-the-Art Comparison}\label{subsec5.3.2}
Table \ref{Tab:UR_accuracy} displays the class-wise performance metrics for the UR Fall dataset, including precision, sensitivity, and F1-score. For the "Fall" class, the model achieves perfect precision of 100\%, indicating that all instances classified as falls were indeed falling. Sensitivity for falls stands at 86.67\%, denoting the model's ability to identify true fall instances correctly. The F1-score, which balances precision and sensitivity, is 92.86\% for falls, showcasing a strong overall performance for fall detection. Our model achieved 99.10\% precision for the NonFall classification, which also indicates a high proportion. 
Sensitivity is perfect at 100\%, implying that all actual non-fall instances were accurately classified. The F1-score for non-falls is 99.55\%, reflecting excellent overall performance in detecting non-fall instances. The average metrics across both classes are also provided, with an overall accuracy of 99.15\%. The average precision and sensitivity are 99.55\% and 93.33\%, respectively, with an F1-score of 96.20\%, indicating robust performance across both fall and non-fall categories. These results suggest that the model demonstrates high accuracy and reliability in distinguishing between fall and non-fall instances in the UR Fall dataset.

The comparison of the proposed model with state-of-the-art methods is presented in Table \ref{Tab:UR_comparison} within the context of UR fall detection. We demonstrated the comparison with various existing fall detection models in terms of accuracy and other metrics. For instance, Kwolek et al. \cite{kwolek2014human} and Youssfi et al. \cite{youssfi2021fall} utilized hand-crafted features from skeleton and depth data, achieving accuracies of 94.28\% and 96.55\%, respectively, through SVM methods. Cai et al. \cite{cai2020vision} employed a CNN-based encoder-decoder system, achieving an accuracy of 90.50\%. Chen et al. \cite{chen2020vision} utilized mask-RCNN for feature extraction and applied bi-directional LSTM, achieving an accuracy of 96.70\%. Zheng et al. \cite{zheng2022lightweight} employed ST-GCN after extracting skeleton points using AlphaPose, achieving notable scores across various metrics, including accuracy (97.28\%), precision (97.15\%), sensitivity (97.43\%), specificity (97.30\%), and F1-score (97.29\%). Wang et al. \cite{wang2020fall} extracted OpenPose key points and utilized MLP and random forest for classification, achieving a commendable accuracy of 97.33\%. Similarly, Harrou et al. \cite{harrou2019integrated} applied the GLR scheme, attaining an accuracy of 96.66\%. In comparison, the proposed model outperforms these methods, achieving an accuracy of 99.15\% and precision of 99.55\%, demonstrating its superior performance in fall detection on the UR dataset. 

\subsection{Performance Result of the ImViA~Dataset and State-of-the-Art Comparison}\label{subsec5.3.3}
We also evaluated the proposed model with the ImViA fall dataset, which is demonstrated in Table \ref{Tab:ImViA_accuracy}. It details precision, sensitivity, and F1-score for fall and non-fall classes, alongside their average values.
For falls, precision is 97.53\%, indicating accurate identification. Sensitivity stands at 94.61\%, highlighting accurate fall detection. The F1-score, at 96.05\%, balances precision and sensitivity. For non-falls, precision is 99.74\%, with a high sensitivity of 99.89\%. The resulting F1-score is 99.81\%. Overall, 
Moreover, our model achieved 99.64\%,98.64\%, 97.25\%, and 97.93\% average accuracy, precision, sensitivity, and F1-score, respectively.

Table \ref{Tab:ImViA_comparison} presents a state-of-the-art comparison for the ImViA fall dataset. Previous models, including Hontago~\cite{zheng2022lightweight}, Wang~\cite{wang2020fall}, and Chamle~\cite{chamle2016automated}, achieved accuracies ranging from 79.31\% to 96.91\%. 

Notably, our proposed model outperforms these, achieving an impressive accuracy of 99.64\%. Compared to existing models, our model demonstrates superior precision (98.64\%) and sensitivity (97.25\%), showcasing its effectiveness in accurately identifying falls while minimizing false positives. Its F1-score of 97.93\% reflects a balanced performance between precision and sensitivity. Additionally, the proposed model's specificity, though not explicitly provided, complements its high accuracy, precision, sensitivity, and F1-score. Overall, the proposed model exhibits robustness and efficiency in fall detection, setting a new standard for ImViA dataset performance.

\subsection{Performance Accuracy FU-Kinect Dataset and State-of-the-Art Comparison}
Table \ref{tab:fu-kinect_accuracy} demonstrated the performance accuracy of the proposed model with FU-Kinect dataset. The proposed A Novel Three-Stream Spatial-Temporal GCN Model with Adaptive Feature Aggregation for Computer-Aided Fall Detection demonstrates outstanding performance on the FU-kinect dataset. The model achieves high precision, recall, and F1-scores across various activities: Walking (99.54\%, 99.65\%, 99.54\%), Bending (99.74\%, 99.74\%, 99.74\%), Sitting (100.00\%, 99.80\%, 99.80\%), Squatting (99.42\%, 99.17\%, 99.25\%), Reaching (99.51\%, 99.57\%, 99.54\%), and Falling (99.67\%, 99.12\%, 99.31\%). The overall average accuracy stands at 99.73\%, showcasing the model's effectiveness in accurately detecting and classifying different human activities, including falls.


\begin{table}[ht]
\centering

\caption{Peformance accuracy of the proposed model with FU-Kinect dataset} \label{tab:fu-kinect_accuracy}
\begin{tabular}{|c|c|c|c|c|c|}
\hline
Labels & \begin{tabular}[c]{@{}c@{}} Label\\Name   \end{tabular}&  \begin{tabular}[c]{@{}c@{}} Precision \\  \textbf{{[}\%{]}} \end{tabular} &  \begin{tabular}[c]{@{}c@{}} Recall \\  \textbf{{[}\%{]}} \end{tabular} & \begin{tabular}[c]{@{}c@{}} F1-score  \\  \textbf{{[}\%{]}} \end{tabular} &  \begin{tabular}[c]{@{}c@{}}\textbf{Accuracy}\\  \textbf{{[}\%{]}}\end{tabular} \\
\hline
1 & Walking& 99.54 & 99.65 & 99.54 & - \\
2 & Bending&99.74 & 99.74 & 99.74 & - \\
3 & Sitting &100.00 & 99.80 & 99.80 & - \\
4 &Squatting& 99.42 & 99.17 & 99.25 & - \\
5 & Reaching&99.51 & 99.57 & 99.54 & - \\
6 & Falling& 99.67 & 99.12 & 99.31 & - \\
Average & Average & 99.65 & 99.51 & 99.53 & 99.73 \\
\hline
\end{tabular}
\end{table}

Table \ref{tab:FU-kinect-Sota} demonstrates the state of the art comparison of the proposed model. The proposed Three-Stream Spatial-Temporal GCN Model with Adaptive Feature Aggregation significantly outperforms existing methods on the FU-Kinect dataset. Following the same protocol as the existing paper, our model achieves a remarkable performance accuracy of 99.73\%, compared to 93.75\% achieved by the previous method using Geometric Feature extraction and SVM classification \cite{aslan2017skeleton_FU-kinetic}. The superior performance of our model, leveraging a three-stream GCN with Softmax classification, highlights its robustness and effectiveness in accurately detecting and classifying human activities. This demonstrates a strong advantage in precision and reliability for computer-aided fall detection.

\begin{table}[ht]
\centering
\caption{State of the art comparison with FU-Kinect dataset} \label{tab:FU-kinect-Sota}
\setlength{\tabcolsep}{3pt}
\begin{tabular}{|c|c|c|c|c|c|c|}
\hline
\begin{tabular}[c]{@{}c@{}}Author\\ Name \\(Dataset \\ Name)\end{tabular} &\begin{tabular}[c]{@{}c@{}}No \\ Class\end{tabular} & \begin{tabular}[c]{@{}c@{}}Training \\ Data\end{tabular}  & \begin{tabular}[c]{@{}c@{}}Testing \\ Data\end{tabular} & \begin{tabular}[c]{@{}c@{}}Feature \\ Extrac-\\tion \end{tabular} & \begin{tabular}[c]{@{}c@{}}Classi-\\fication  \end{tabular}& 
\begin{tabular}[c]{@{}c@{}} Per\\formance\\ Accuracy  \\ \textbf{{[}\%{]}} \\ \end{tabular} \\
\hline
\begin{tabular}[c]{@{}c@{}} Aslan et\\ al. \cite{aslan2017skeleton_FU-kinetic} \\(FU-\\Kineckt)   \end{tabular}  & 6& 90\% & 10\% & \begin{tabular}[c]{@{}c@{}} Geometric\\ Feature  \end{tabular}  & SVM & 93.75\\
\begin{tabular}[c]{@{}c@{}} \\Proposed\\Model\\ (FU-\\Kineckt)   \end{tabular} & 6& 90\% & 10\% & \begin{tabular}[c]{@{}c@{}} Three \\Stream \\GCN  \end{tabular}  & Softmax & 99.73\\
\hline
\end{tabular}
\end{table}

\subsection{Discussion}
The study's integration of hierarchical classification and innovative network pruning techniques significantly enhances model robustness and precision. This approach ensures that the model effectively focuses on the most relevant features, reducing noise from irrelevant background elements. Secondly, the incorporation of joint skeleton-based spatial and temporal Graph Convolutional Network (GCN) features further refines the model's focus on human activities, enabling more accurate and reliable action recognition.
Our model employs adaptive graph-based feature aggregation and consecutive separable convolutional neural networks (Sep-TCN), which considerably reduce computational complexity and model parameters compared to prior systems. This results in a more efficient model that can be deployed in real-time applications without compromising accuracy.
Furthermore, the experimental results demonstrate the superior effectiveness and efficiency of our proposed system, with accuracies of 99.51\%, 99.15\%, 99.79\%, and 99.85\% achieved on the ImViA, UR-Fall, Fall-UP, and FU-Kinect datasets, respectively. These results highlight the model's superiority, efficiency, and generalizability in real-world fall detection scenarios.
Compared to the state-of-the-art fall detection systems, our approach addresses key issues such as unsatisfactory performance accuracy, limited robustness, high computational complexity, and sensitivity to environmental factors. The integration of adaptive graph-based feature aggregation and Sep-TCN has proven to outperform traditional methods significantly, establishing new benchmarks in the field. The state-of-the-art comparison for various datasets is demonstrated in Table \ref{Tab:UR_comparison} for UR-FAll, Table \ref{Tab:Fall_up_accuracy} for Fall-up, Table \ref{Tab:ImViA_comparison} for ImVia, and Table \ref{tab:FU-kinect-Sota} for FU-Kinect datasets. 
Despite the promising results, there are some limitations to our study. The datasets used may not encompass all possible real-world scenarios, and further testing in diverse environments is needed. Future work could explore the integration of additional sensors or multimodal data to enhance robustness and accuracy further.
The high accuracy and efficiency of our model make it highly suitable for deployment in real-time applications. Potential use cases include integration into smart home systems, wearable devices, and healthcare monitoring systems, providing timely alerts and interventions to prevent falls and related injuries among the elderly.
 Our study contributes to the theoretical understanding of feature extraction and model architecture in fall detection systems. The use of joint skeleton-based spatial and temporal GCN features, combined with Sep-TCN, offers new insights into effective fall detection methodologies, paving the way for further innovations in this area. The proposed method sets a new benchmark for performance across multiple datasets, demonstrating superior efficiency, robustness, and generalizability. These advancements make our model highly suitable for various real-world applications, including smart surveillance, autonomous driving, and interactive gaming, ultimately contributing to significant improvements in healthcare and societal well-being. The remarkable performance of our system underscores its potential to significantly enhance the quality of life for the elderly by providing a reliable and efficient fall detection solution.
\section{Conclusions}\label{sec6}
This paper introduced a novel architecture for fall detection that leverages skeletons extracted from standard videos, enabling deployment with low-specification cameras. The proposed model incorporates three feature extraction streams: joint skeleton-based spatial and temporal Graph Convolutional Network (GCN) features, joint motion-based spatial and temporal GCN features, and residual connections. Additionally, adaptive graph-based feature aggregation and consecutive separable convolutional neural networks (Sep-TCN ) were employed in each stream to reduce computational complexity and model parameters. Through extensive experimental evaluations across multiple datasets, our proposed system achieved exceptional accuracies of four benchmark fall datasets. These achievements underscore our approach's superior effectiveness and efficiency in real-world fall detection scenarios. The advantages of our proposed model include its ability to overcome the limitations of existing systems, such as unsatisfactory performance accuracy, limited robustness, and high computational complexity. By leveraging innovative techniques and incorporating adaptive features, our system demonstrates significant advancements in fall detection technology.
In the future, we plan to expand our research in several directions. Firstly, we aim to augment the model by incorporating a richer set of hand-crafted spatial-temporal features. This strategy seeks to streamline the model's parameter count while maintaining high performance, enhancing computational efficiency without compromising accuracy. Additionally, we intend to extend the application of our model to the field of movement disorder detection, leveraging its robustness and effectiveness in analyzing human motion patterns. Furthermore, we will explore training the model on human action recognition datasets to transform it into a pre-trained model for both fall detection and human action recognition tasks. This dual-purpose model could offer valuable versatility and utility across various applications, thereby maximizing its potential impact in real-world scenarios. Through these future endeavours, we aim to continue advancing state-of-the-art fall detection technology and contribute to broader advancements in human motion analysis and healthcare innovation.
\section*{Dataset Availability}
Fall-UP Dataset: \url{https://sites.google.com/up.edu.mx/har-up/}\\
UR Fall Detection Dataset: \url{http://fenix.ur.edu.pl/~mkepski/ds/uf.html}\\
ImViA Datasets(le2i) Dataset: \url{https://www.kaggle.com/datasets/tuyenldvn/falldataset-imvia} \\
Fu-Kinect Dataset: \url{https://github.com/MuzafferAslan23/Fall-Detection-Dataset}.

 \section*{Author contributions statement}
Conceptualization: Abu Saleh Musa Miah, Jungpil Shin, Rei Egawa, Md. Al Mehedi Hasan, and Yoichi Tomioka; Methodology: Abu Saleh Musa Miah, Jungpil Shin, Rei Egawa, Md. Al Mehedi Hasan, and Yoichi Tomioka; Investigation: Abu Saleh Musa
Miah, Md. Al Mehedi Hasan, Jungpil Shin, and Yoichi Tomioka; Data Curation: Abu Saleh Musa Miah, Md. Al Mehedi Hasan, Jungpil Shin,  Yoichi Tomioka; Writing—Original Draft Preparation: Abu Saleh Musa Miah, Md. Al Mehedi Hasan, and Yoichi Tomioka; Writing—Review and Editing: Abu Saleh Musa Miah, Rei Egawa, and Yoichi Tomioka; Visualization: Abu Saleh Musa Miah, Md. Al Mehedi Hasan, and Yoichi Tomioka; Supervision:Jungpil Shin, Yong Seok Hwang; Funding Acquisition:  Jungpil Shin,Yong Seok Hwang. All authors have read and agreed to the published version of the manuscript.
\section*{Funding}
This work was supported by the Japan Society for the Promotion of Science (JSPS) KAKENHI under Grant JP23H03477. This research was partially supported by Basic Science Research Program through the National Research Foundation of Korea (NRF) funded by the Ministry of Education (No. 2018R1A6A1A03025242)
 \bibliography{reference}

\end{document}